\documentclass[letterpaper, 10 pt, conference]{ieeeconf}  

\IEEEoverridecommandlockouts                              

\usepackage{times}
\usepackage{epsfig}
\usepackage{graphicx}
\usepackage{amsmath}
\usepackage{bm}
\usepackage{amssymb}
\usepackage{tabularx}
\usepackage{amsmath}
\usepackage{siunitx}
\usepackage{booktabs}
\usepackage{multirow}
\usepackage[table,xcdraw]{xcolor}
\usepackage{float}
\usepackage[normalem]{ulem}
\useunder{\uline}{\ul}{}
\newcommand{\etal}{et al.}
\usepackage{cite}
\newcommand{\minus}{\scalebox{0.5}[1.0]{$-$}}
\usepackage[pagebackref=true,breaklinks=true,colorlinks,bookmarks=false]{hyperref}

\title{\LARGE \bf Multimodal Scale Consistency and Awareness for Monocular Self-Supervised Depth Estimation}

\author{Hemang Chawla$^*$, Arnav Varma$^*$, Elahe Arani, and Bahram Zonooz
\thanks{$^*$Equal Contribution. All authors are with the Advanced Research Lab, Navinfo Europe, The Netherlands
{\tt\small \{hemang.chawla, arnav.varma, elahe.arani, b.yoosefizonooz\}@navinfo.eu}}%
}

\begin{document}

\onecolumn
{
\noindent{
\large
\textbf{This paper has been accepted for publication in the proceedings of the 
\textit{2021 IEEE International Conference on Robotics and Automation (ICRA)}, May 30 - June 5, 2021, Xi'an, China}}

\bigskip\bigskip
\noindent{
\large
IEEE Copyright notice:\\\\
\normalsize
\copyright 2021 IEEE. Personal use of this material is permitted. Permission from IEEE must be obtained for all other uses, in any current or future media, including reprinting /republishing this material for advertising or promotional purposes, creating new collective  works,  for  resale  or  redistribution  to  servers  or  lists,  or  reuse  of  any  copyrighted  component  of  this  work  in  other works.}

\bigskip\bigskip
\noindent{
\large
Cite as:\\\\
\noindent\fbox{%
    \parbox{\textwidth}{%
    \noindent
    \normalsize
       H. Chawla, A. Varma, E. Arani, and B. Zonooz, ``Multimodal Scale Consistency and Awareness for Monocular Self-Supervised Depth Estimation," \textit{2021 IEEE International Conference on Robotics and Automation (ICRA), Xi'an, China, IEEE (in press), 2021.}%
    }%
}}

\bigskip\bigskip
\noindent{
\large
\textsc{Bib}\TeX:\\\\
\noindent\fbox{%
    \parbox{\textwidth}{%
    \noindent
    \normalsize
    \texttt{\noindent @inproceedings\{chawlavarma2021multimodal,\\
        author=\{H. \{Chawla\} and A. \{Varma\} and E. \{Arani\} and B. \{Zonooz\}\},\\
        booktitle=\{2021 IEEE International Conference on Robotics and Automation (ICRA)\}, \\
        title=\{Multimodal Scale Consistency and Awareness for Monocular Self-Supervised Depth Estimation\}, \\
        location=\{Xi'an, China\},\\
        publisher=\{IEEE (in press)\},\\
        year=\{2021\}\}     
        }%
    }%
}}
}
\normalsize
\twocolumn

\maketitle
\thispagestyle{empty}
\pagestyle{empty}

\begin{abstract}
Dense depth estimation is essential to scene-understanding for autonomous driving. However, recent self-supervised approaches on monocular videos suffer from scale-inconsistency across long sequences. Utilizing data from the ubiquitously copresent global positioning systems (GPS), we tackle this challenge by proposing a dynamically-weighted \textit{ GPS-to-Scale (g2s)} loss to complement the appearance-based losses. We emphasize that the GPS is needed only during the multimodal training, and not at inference. 
The relative distance between frames captured through the GPS provides a scale signal that is independent of the camera setup and scene distribution, resulting in richer learned feature representations. Through extensive evaluation on multiple datasets, we demonstrate scale-consistent and -aware depth estimation during inference, improving the performance even when training with low-frequency GPS data.
\end{abstract}
\section{Introduction}
\label{sec:introduction}
Robots and autonomous driving systems require scene-understanding for planning and navigation. Therefore, spatial perception through depth estimation is essential for enabling complex behaviors in unconstrained environments. 
Even though sensors such as LiDARs can perceive depth at metric-scale~\cite{barsan2018learning}, their output is sparse and they are expensive to use.
In contrast, \textit{monocular} color cameras are compact, low-cost, and consume less energy. While traditional camera-based approaches rely upon hand-crafted features from multiple views~\cite{karsch2012depth}, deep learning based approaches can predict depth from a single image. Among these, self-supervised methods that predict the ego-motion and depth simultaneously by view-synthesis of adjacent frames~\cite{casser2019depth,godard2019digging, zhou2017unsupervised} are preferred over supervised methods that require accurate ground truth labels for training~\cite{fu2018deep,tang2018ba,zhou2018deeptam}.
\begin{figure}[t]
\centering
\includegraphics[width=\linewidth]{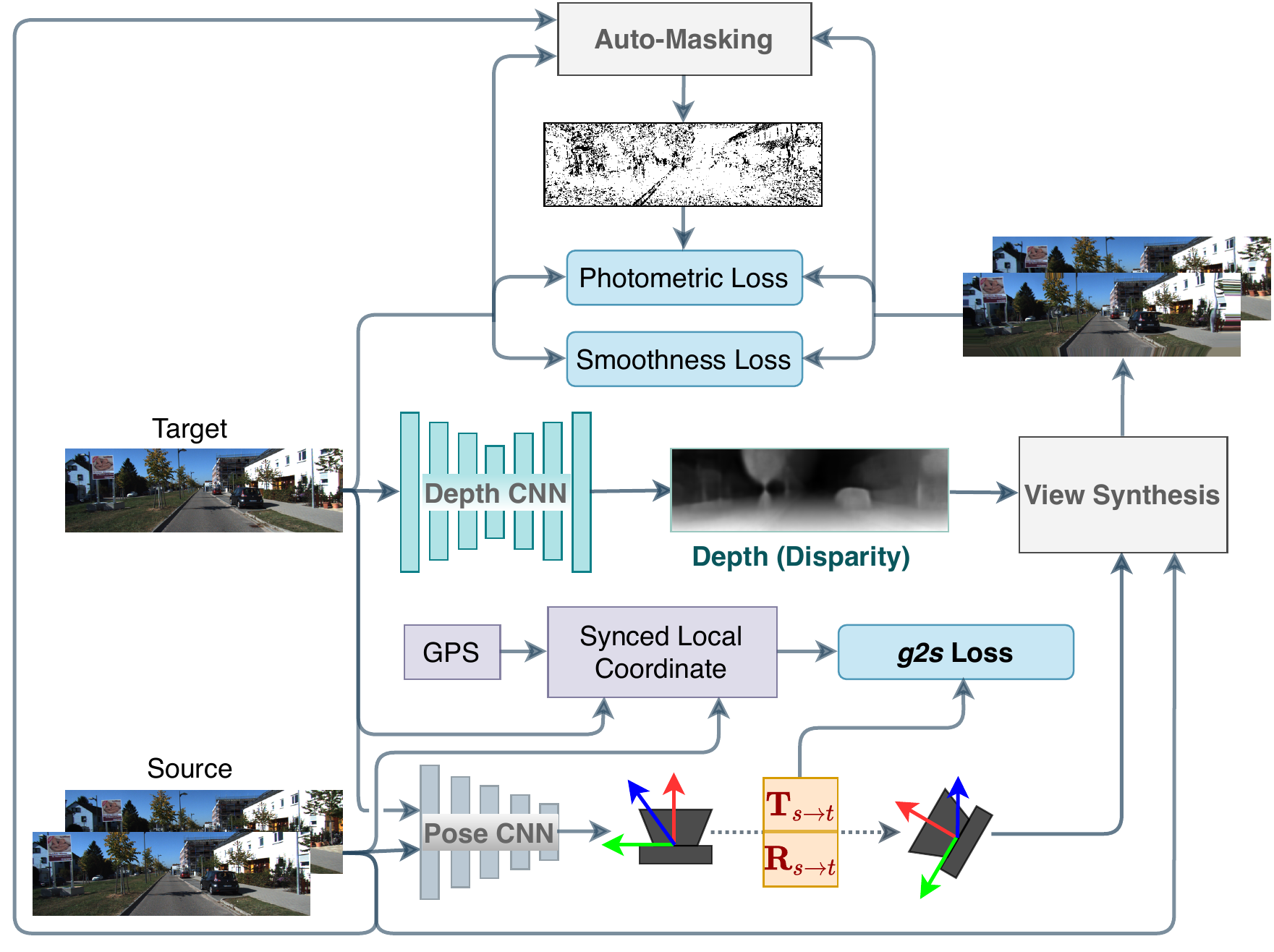}
\caption{A schematic of our proposed multimodal self-supervised depth and ego-motion prediction network for monocular videos. We introduce a \textit{GPS-to-Scale (g2s)} loss that leads to scale-consistent and -aware estimates during inference.}
\label{fig:schematic}
\end{figure}



However, monocular vision inherently suffers from scale ambiguity. Additionally, the self-supervised approaches introduce scale-inconsistency in estimated depth across different video snippets~\cite{bian2019unsupervised}. 
Consequently, most of the existing methods scale the estimated relative depth using the LiDAR ground truth during evaluation. Recent methods tackling this problem utilize additional 3D geometric constraints to introduce scale-consistency~\cite{bian2019unsupervised, mahjourian2018unsupervised}, but require at least some depth or stereo supervision to predict at metric-scale~\cite{guizilini2020robust, roussel2019monocular}. Nevertheless, obtaining metric scale predictions at low cost is necessary for practical deployment.%

Since self-supervised learning allows training on large and varied data including crowdsourced data~\cite{chawla2020crowdsourced, gordon2019depth}, the ubiquitous GPS copresent with videos can be employed for multimodal training.
Taking cues from how cross-modal learning leads to richer learned feature representations~\cite{ngiam2011multimodal, ramachandram2017deep}, 
we hypothesize that the relative distance between image frames captured from the GPS can provide a scale signal that complements commonly used appearance-based losses to predict scale-consistent and -aware improved estimates.





In this work, we propose a \textit{GPS-to-Scale (g2s)} loss that utilizes the ratio of magnitudes of the relative translation measured by the GPS and the relative translation predicted by the pose network to enforce scale-consistency and -awareness on the depth predictions, linked together via the perspective projection model~\cite{zhou2017unsupervised}. Scale consistency implies that the standard deviation of the depth scale factors across the video is low. Scale awareness implies that the mean scale factor is close to $1$. Note that this GPS information is only used during the training, while the inference is directly performed on the unlabeled monocular videos. Furthermore, we compare different weighting strategies for the proposed loss and demonstrate that exponentially increasing the weight on \textit{g2s} over the epochs leads to the best performance.  Experiments on the KITTI raw~\cite{geiger2013vision} Eigen~\cite{eigen2014depth} split as well as the improved KITTI depth benchmark~\cite{Uhrig2017THREEDV} show that adding the \textit{g2s} loss improves performance and scale-consistency over state-of-the-art-methods,
even with low-frequency planar GPS (without altitude). 
Finally, with experiments on out-of-distribution Make3D~\cite{saxena2008make3d} and Cityscapes~\cite{cordts2016cityscapes} datasets, we show that the introduced scale-consistency and -awareness is present across domains in comparison with other methods. 

\section{Related Work}
\label{sec:related}
Estimating scene depth is a long-standing problem in computer vision. 
Traditional approaches solve this by utilizing disparity across multiple views within a non-linear optimization framework~\cite{karsch2012depth, mur2015orb}. Supervised methods that produce high-quality estimates have also been proposed~\cite{fu2018deep, tang2018ba, zhou2018deeptam}, but necessitate the availability of accurate ground truth and cross-calibration of sensors for training. Instead, using view-synthesis as a signal, self-supervised methods produce accurate depth maps from stereo image pairs~\cite{garg2016unsupervised, godard2017unsupervised} or monocular video snippets~\cite{casser2019depth,godard2019digging, zhou2017unsupervised}. We focus on methods employing purely monocular setups, as they are more pervasive and do not depend upon prior knowledge of relative rotation and translation of the stereo camera pairs. 
However, most existing monocular approaches utilize only appearance-based losses with the assumption of brightness consistency that limits training on small video subsequences without any long sequence constraints. Hence, the depth and ego-motion estimates from these methods suffer from scale-inconsistency along with the global scale-ambiguity present in monocular vision. Therefore, ground truth LiDAR depth maps~\cite{godard2019digging} or camera height~\cite{xue2020toward} are used during inference to recover per-image scale. 

Methods addressing this problem add 3D-geometry-based losses to introduce scale-consistency~\cite{bian2019unsupervised,mahjourian2018unsupervised}, yet utilize at least some depth or stereo supervision to introduce scale-awareness ~\cite{guizilini2020robust, roussel2019monocular}.
Recently~\cite{guizilini20203d} introduced a similar instantaneous velocity based multi-modal supervision. However, access to instantaneous velocity may require the use of inertial measurement units (IMU) that are less ubiquitous. In contrast, GPS is often copresent, such as in dashboard cameras albeit with lower frequency, allowing training on more data.
In this work, we introduce a \textit{GPS-to-scale (g2s)} loss that produces improved scale-consistent and -aware results even with low-frequency planar GPS without altitude. 

\section{Method}
\label{sec:method}
Our objective is to simultaneously train depth and ego-motion prediction networks that produce scale-consistent and -aware estimates from only a monocular color camera during inference. Here we describe the baseline network and appearance-based losses for self-supervised learning, followed by the motivation and description of our proposed dynamically-weighted \textit{GPS-to-Scale (g2s)} loss.

\subsection{Overview}
Given a set of $n$ images from a video sequence, and $m$ loosely corresponding GPS coordinates, the inputs to the networks are a sequence of temporally consecutive RGB image triplets $\{I_{\minus 1}, I_0, I_1\} \in \mathbb{R}^{H\times W \times 3}$ and the the synced GPS coordinates $\{ G_{\minus 1}, G_0, G_1\} \in \mathbb{R}^{3}$, when available. The depth network, $f_D:\mathbb{R}^{H\times W \times 3} \rightarrow \mathbb{R}^{H \times W}$, outputs dense depth (or disparity) for each pixel coordinate $p$ of a single image. Simultaneously, the ego-motion network, $f_{E}:\mathbb{R}^{2 \times H\times W \times 3} \rightarrow \mathbb{R}^6$, outputs relative translation $(t_x, t_y, t_z)$ and rotation $(r_x, r_y, r_z)$  forming the affine transformation  \tiny $\begin{bmatrix} \hat{{R}} & \hat{{T}} \\ {0} & 1 \end{bmatrix}$ \normalsize $\in \text{SE(3)}$ between a pair of adjacent images.  
The predicted depth $\hat{{D}}$ and ego-motion $\hat{{T}}$ are linked together via the perspective projection model~\cite{zhou2017unsupervised},
that warps the source ($s$) images $I_s \in \{I_{\minus 1}, I_1\}$ to the target ($t$) image $ I_t \in \{I_0\}$, given the camera intrinsics $K$. 

We establish a strong baseline by following the best practices of appearance-based learning from Monodepth2~\cite{godard2019digging}.
The networks are trained using the appearance-based \textit{photometric} loss between the real and synthesized target images, as well as a \textit{smoothness} loss for depth regularization in low texture scenes~\cite{godard2019digging}. Following ~\cite{li2018monocular, godard2019digging}, we use auto-masking (M) to disregard the temporally stationary pixels in the image triplets. The total appearance-based loss is calculated by upscaling the predicted depths from intermediate decoder layers to the input resolution.

Additionally, we introduce the dynamically-weighted \textit{g2s} loss that enforces scale-consistency and -awareness using the ratio of the measured and estimated translation magnitudes. Fig. \ref{fig:schematic} illustrates the complete architecture that uses the proposed method.

 \subsection{GPS-to-Scale (g2s) Loss}
 \label{sec:g2s_loss}
 Appearance-based losses provide supervisory signals on short monocular subsequences. This leads to scale-inconsistency of the predictions across long videos. 
Approaches addressing this problem through 3D-geometry-based losses provide a signal that depends upon the camera setup and the scene distribution~\cite{bian2019unsupervised, mahjourian2018unsupervised}. Therefore, we introduce the \textit{GPS-to-Scale (g2s)} loss that provides an independent cross-modal signal leading to scale-consistent and -aware estimates. 
 
\textbf{Synced Local Coordinates}: The GPS information, ubiquitously copresent with videos, consists of the latitude, longitude, and optionally the altitude of the vehicle. First, we convert these geodetic coordinates to local coordinates 
$\bm{G} = \{x_g, y_g, z_g\}$ using the Mercator projection such that,
\begin{align}
\centering
x_g &= \cos \left(\dfrac{\pi \cdot \text{lat}_0}{180}\right) r_{e} \log \left(\tan\dfrac{\pi \cdot (90 + \text{lat})}{360}\right)\\
y_g &= \text{alt}\\
z_g &= \cos \left(\dfrac{\pi \cdot \text{lat}_0}{180}\right) r_{e} \dfrac{\pi \cdot \text{lon}
}{180}
\end{align}
where $r_{e} = \SI{6378137}{\m}$ is taken as the radius of earth.
Since the GPS frequency may be different from the frame-rate of the captured video, we additionally sync these local coordinates with the images using their respective timestamps. 

Utilizing the ratio of the relative distance measured by the GPS and the relative distance predicted by the network, we additionally impose our proposed \textit{g2s} loss given by,
\begin{equation}
\label{eq:g2s_loss}
    \mathcal{L}_{g2s} = \sum\limits_{s, t} 
    \left(\dfrac{\lVert {G}_{s \rightarrow t} \rVert_2}{\lVert \hat{{T}}_{s\rightarrow t} \rVert_2} - 1 \right)^2
\end{equation}
where $s \in \{-1,1\}$ and $t \in \{0\}$. Following~\cite{zhou2017unsupervised} we remove static frames while training, thereby allowing the \textit{g2s} loss to be differentiable for all plausible inputs.

\textbf{GPS noise and bias}: By forming this loss upon the translation magnitude instead of the individual components $(t_x, t_y, t_z)$, we account for any noise or systemic bias that may be present in the GPS measurements~\cite{das2018experimental}.
This loss encourages the ego-motion estimates to be closer to the common metric scale across the image triplets, thereby introducing the scale-consistency and -awareness which is extended to the depth estimates that are tied to the ego-motion via the perspective projection model.


Note that CNNs tend to learn surface statistical regularities by exploiting superficial clues (or shortcuts)  specific to the distribution being trained on~\cite{jo2017measuring, geirhos2020shortcut}. 
Since the GPS signal does not depend upon the specific scene distribution or camera setup, we hypothesize that adding our proposed \textit{g2s} loss in a multimodal context can help to disentangle intended higher-level abstractions ~\cite{ramachandram2017deep} from the shortcut features to improve the estimates and help in generalizing scale-consistency to out-of-distribution (o.o.d.) datasets. 


\subsection{Dynamic Weighting Strategy}
\label{sec:method_weighting_strategy}
The networks learn to synthesize more plausible views of the target images $I_t$ by improving their depth and ego-motion predictions over the training epochs. 
Thus, heavily penalizing the networks for the incorrect scales during the early training can interfere with the learning of individual translations, rotations, and pixel-wise depths.
Hence, we dynamically weigh the \textit{g2s} loss in an exponential manner to provide a scale signal  that is low in the beginning and increases as the training progresses. The weight $w$ to the \textit{g2s} loss $\mathcal{L}_{g2s}$ is given by,
\begin{equation}
\label{eq:weighting}
    w = \exp{\left(\text{epoch} - \text{epoch}_{\text{max}}\right)}.
\end{equation}

 
\subsection{Final Training Loss}
The final loss combining the appearance-based losses~\cite{zhou2017unsupervised, godard2019digging} with Eqs. \ref{eq:g2s_loss} and \ref{eq:weighting} is given by,
\begin{equation}
\label{eq:final_loss}
\mathcal{L} = \mathcal{L}_{\text{appearance}} + w \cdot \mathcal{L}_{g2s},
\end{equation}
which is averaged over each batch. 
\section{Experiments}
\label{sec:experiments}
For all our experiments, we follow the setup of Monodepth2~\cite{godard2019digging}.

\begin{figure*}[t]
\centering
  \includegraphics[width=\linewidth]{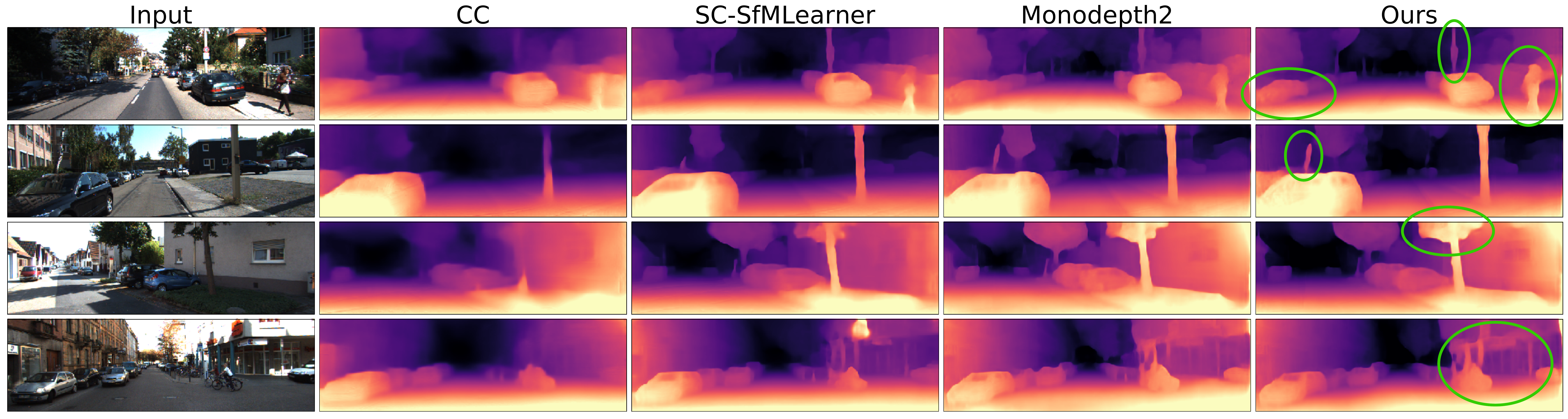}
  \caption{Single-image depth estimates on the KITTI Eigen split. Our method produces sharper, high-quality predictions that preserve more structure when compared against existing methods.} 
\label{fig:kitti_maps}
\end{figure*}

\begin{table*}[tb]
\centering
\caption{\textit{Per-image scaled} dense depth prediction (without post-processing) on KITTI \textit{Original}~\cite{geiger2013vision} and \textit{Improved}~\cite{Uhrig2017THREEDV}.}
\resizebox{0.85\textwidth}{!}{
\begin{tabular}{|c|l|c|cccc|ccc|}
\hline
\multirow{2}{*}{GT}  &\multirow{2}{*}{\textbf{Methods}} & \multirow{2}{*}{\textbf{Resolution}} & \multicolumn{4}{c|}{\textbf{\cellcolor{red!25}Error$\downarrow$}} & \multicolumn{3}{c|}{\textbf{\cellcolor{blue!25}Accuracy$\uparrow$}} \\ \cline{4-10}
 &    &  & Abs Rel & Sq Rel & RMSE & RMSE log & $\delta<1.25$ & $\delta<1.25^2$ & $\delta<1.25^3$ \\ \hline \hline
\multirow{10}{*}{\rotatebox[origin=c]{90}{Original}}
 & SfMLearner\cite{zhou2017unsupervised} & 416$\times$128 & 0.208 & 1.768 & 6.856 & 0.283 & 0.678 & 0.885 & 0.957 \\
  & GeoNet~\cite{yin2018geonet} & 416$\times$128 & 0.155 & 1.296 & 5.857 & 0.233 & 0.793 & 0.931 & 0.973 \\
  & Vid2Depth~\cite{mahjourian2018unsupervised} & 416$\times$128 & 0.163 & 1.240 & 6.220 & 0.250 & 0.762 & 0.916 & 0.968 \\
  & Struct2Depth~\cite{casser2019depth} & 416$\times$128 & 0.141 & 1.026 & 5.291 & 0.215 & 0.816 & 0.945 & 0.979 \\
  & VITW~\cite{gordon2019depth} & 416$\times$128 & 0.128 & 0.959 & 5.230 & 0.212 & 0.845 & 0.947 & 0.976 \\
  & Roussel \etal~\cite{roussel2019monocular} & 416$\times$128 & 0.179 & 1.545 & 6.765 & 0.268 & 0754 & 0916 & 0.966 \\
  & CC~\cite{ranjan2019competitive} & 832$\times$256 & 0.140 & 1.070 & 5.326 & 0.217 & 0.826 & 0.941 & 0.975 \\
  & SC-SfMLearner~\cite{bian2019unsupervised} & 832$\times$256 & 0.137 & 1.089 & 5.439 & 0.217 & 0.830 & 0.942 & 0.975 \\
  & Monodepth2~\cite{godard2019digging} & 640$\times$192 & \underline{0.115} & \underline{0.903} & 4.863 & \underline{0.193} & \textbf{0.877} & \textbf{0.959} & \textbf{0.981} \\
  & SG Depth~\cite{klingner2020selfsupervised} & 640$\times$192 & 0.117 & 0.907 & \textbf{4.844} & 0.194 & \underline{0.875} & \underline{0.958} & \underline{0.980} \\
  & Ours & 640$\times$192 & \textbf{0.112} & \textbf{0.894} & \underline{4.852} & \textbf{0.192} & \textbf{0.877} & \underline{0.958} & \textbf{0.981} \\
  \hline
\multirow{5}{*}{\rotatebox[origin=c]{90}{Improved}} 
 & SfMLearner$^*$~\cite{zhou2017unsupervised} & 416$\times$128 & 0.176 & 1.532 & 6.129 & 0.244 & 0.758 & 0.921 & 0.971 \\
  & Geonet$^*$~\cite{yin2018geonet} & 416$\times$128 & 0.132 & 0.994 & 5.240 & 0.193 & 0.883 & 0.953 & 0.985 \\
  & Vid2Depth$^*$~\cite{mahjourian2018unsupervised}& 416$\times$128 & 0.134 & 0.983 & 5.501 & 0.203 & 0.827 & 0.944 & 0.981 \\
  & Monodepth2~\cite{gordon2019depth} & 640$\times$192 & \underline{0.090} & \textbf{0.545} & \textbf{3.942} & \textbf{0.137} & \textbf{0.914} & \textbf{0.983} & \textbf{0.995} \\
  & Ours & 640$\times$192 & \textbf{0.088} & \underline{0.554} & \underline{3.968} & \textbf{0.137} & \underline{0.913} & \underline{0.981} & \textbf{0.995} \\ \hline
\end{tabular}}

\label{tab:kitti_depth_scaled}
\end{table*}


\begin{figure}[t]
\centering
  \includegraphics[width=\linewidth]{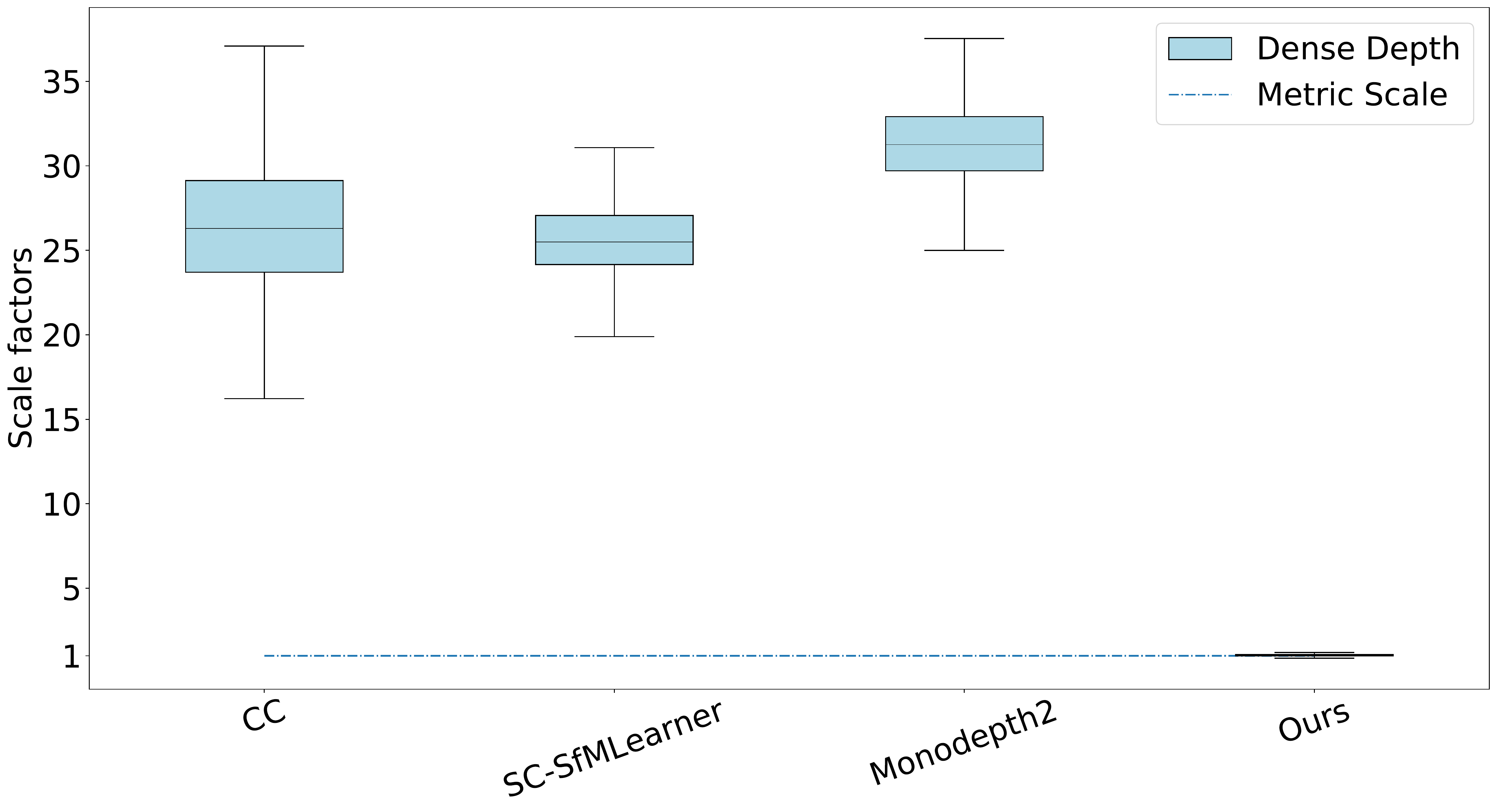}
  \caption{Box-plot visualizing the mean and standard deviation of scale factors for per-image dense depth estimation on the test set of Eigen split \cite{eigen2014depth}. 
  Existing methods scaled the estimated depth using the per-image ground truth during inference. Our method is scale-consistent and -aware and does not need ground truth during inference. }
\label{fig:scale-consistency-boxplot}
\end{figure}
\subsection{Depth Estimation}
Following the established protocols, we compare our depth predictions on the Eigen Split~\cite{eigen2014depth} of KITTI~\cite{geiger2013vision} raw dataset as shown in Tables \ref{tab:kitti_depth_scaled} and \ref{tab:kitti_depth_unscaled}. This contains $39,810$ training
and $697$ test images respectively. The depth is evaluated using metrics from~\cite{eigen2014depth} up to the fixed range of \SI{80}{m}, unless specified otherwise. 
We also evaluate against the \textit{Improved} ground truth depth~\cite{Uhrig2017THREEDV} which contains $652$ ($93$\%) of the $697$ \textit{Original} test images. 
Best results for each metric are in bold. The second best results are underlined. * denotes results when trained on Cityscapes along with KITTI.



\subsubsection{Performance and Scale-Consistency}
\label{sec:expt-scale-consistency}
For evaluating the performance and scale-consistency of depth estimation, we follow the standard procedure of scaling the per-image estimated depths $\hat{D}$ with individual scale factors given by the ratio of the median ground truth depths from LiDAR and the median predicted depths~\cite{zhou2017unsupervised}. A lower standard deviation of the scale factors corresponds to a higher scale-consistency. 

As shown in Table \ref{tab:kitti_depth_scaled}, we outperform existing depth estimation methods on the KITTI \textit{Original} as well as \textit{Improved} ground truths for the Eigen split. This improvement can be attributed to the richer learned feature representations as explained in  Sec \ref{sec:g2s_loss}.
Furthermore, Fig. \ref{fig:kitti_maps} validates our results visually, and demonstrates that the learning of richer feature representation with our proposed multi-modal training leads to sharper depth estimates with improved structure preservation. As discussed earlier in Sec \ref{sec:g2s_loss}, this can be explained by the disentangling of the intended higher-level abstractions from the shortcut features~\cite{ramachandram2017deep,jo2017measuring,geirhos2020shortcut}. 

We also compare the variation of the scale factor 
for different methods as shown in Fig. \ref{fig:scale-consistency-boxplot}. Note that the standard deviation of depth scale factors is the lowest for our method at $0.07$. Unlike previous methods that measure scale-consistency by the standard deviation of the scales normalized by the median scale, we report un-normalized standard deviation.  This shows that the network is able to estimate scale-consistent depths with the use of our proposed \textit{g2s} loss during training.

\subsubsection{Scale-Awareness}
\label{sec:expt-scale-awareness}
We also compare the \textit{unscaled} depth estimates in Table \ref{tab:kitti_depth_unscaled} (LR and HR denote methods trained on low and high resolution  images respectively. pp~\cite{godard2019digging} denotes post-processing during inference). 
As shown, most state-of-the-art monocular self-supervised methods produce poor estimates without the per-image scaling based on the LiDAR ground truth depths. However, our \textit{unscaled} estimates are close to that from Table \ref{tab:kitti_depth_scaled}. As shown in Fig. \ref{fig:scale-consistency-boxplot}, our mean depth scale factor is $\approx 1$  (specifically 1.03), establishing the scale-awareness introduced by our method for monocular depth estimation. 
We outperform Roussel \etal~\cite{roussel2019monocular} which uses stereo pre-training on CityScapes to predict scale-aware monocular depth for KITTI. We also show comparable performance against  Packnet-SfM~\cite{guizilini20203d}
which uses a much heavier depth-estimation-dedicated architecture unlike the ResNet family based methods such as ours. Moreover, while our method has an inference time of \SI{40}{ms} on an NVIDIA 1080Ti GPU, ~\cite{guizilini20203d} has is slower with an inference time of \SI{60}{ms} even on a Titan V100 GPU.   

Hence, through experiments in Sec \ref{sec:expt-scale-consistency} and \ref{sec:expt-scale-awareness}, we demonstrate that the \textit{G2S} loss provides a scale-signal based on relative distance between image frames  resulting in scale-consistent and -aware estimates.

\begin{table*}[tb]

\centering
\caption{\textit{Unscaled} dense depth prediction on KITTI \textit{Original}~\cite{geiger2013vision}.}

\resizebox{0.85\textwidth}{!}{
\begin{tabular}{|c|l|c|cccc|ccc|}
\hline
\multirow{2}{*}{}  &\multirow{2}{*}{\textbf{Methods}} & \multirow{2}{*}{\textbf{Resolution}} & \multicolumn{4}{c|}{\cellcolor{red!25}\textbf{Error$\downarrow$}} & \multicolumn{3}{c|}{\textbf{\cellcolor{blue!25}Accuracy$\uparrow$}} \\ \cline{4-10}
 &    &  & Abs Rel & Sq Rel & RMSE & RMSE log & $\delta<1.25$ & $\delta<1.25^2$ & $\delta<1.25^3$ \\ \hline \hline
 \multirow{7}{*}{\rotatebox[origin=c]{90}{LR}} &  SfMLearner\cite{zhou2017unsupervised} & 416$\times$128 & 0.977 & 15.161 & 19.189 & 3.832 & 0.0 & 0.0 & 0.0 \\
 & Roussel \etal~\cite{roussel2019monocular} & 416$\times$128 & 0.175 & 1.585 & 6.901 & 0.281 & 0.751 & 0.905 & 0.959 \\
 &  CC~\cite{ranjan2019competitive} & 832$\times$256 & 0.961 & 14.672 & 18.838 & 3.280 & 0.0 & 0.0 & 0.0 \\
 &  SC-SfMLearner~\cite{bian2019unsupervised} & 832$\times$256 & 0.961 & 14.915 & 19.089 & 3.264 & 0.0 & 0.0 & 0.0 \\
  &  Monodepth2~\cite{godard2019digging} & 640$\times$192 & 0.969 & 15.126 & 19.199 & 3.489 & 0.0 & 0.0 & 0.0 \\
  &  Packnet-SfM~\cite{guizilini20203d} & 640$\times$192 & \underline{0.111} & \textbf{0.829} & \textbf{4.788} & \underline{0.199} & \underline{0.864} & \textbf{0.954} & \textbf{0.980} \\
 &  Ours  & 640$\times$192 & \underline{0.111} & 0.900 & 4.935 & 0.200 & 0.863 & \underline{0.953} & \underline{0.979} \\
 & Ours (pp) & 640$\times$192 & \textbf{0.109} & \underline{0.860} & \underline{4.855} & \textbf{0.198} & \textbf{0.865} & \textbf{0.954} & \textbf{0.980} \\
\hline
\multirow{3}{*}{\rotatebox[origin=c]{90}{HR}}  & Packnet-SfM~\cite{guizilini20203d} & 1280$\times$384 & \textbf{0.107} & \textbf{0.803} & \textbf{4.566} & 0.197 & \textbf{0.876} & 0.957 & 0.979 \\
 & Ours & 1024$\times$384 & \underline{0.109} & 0.844 & 4.774 & \underline{0.194} & \underline{0.869} & \underline{0.958} & \underline{0.981} \\ 
  & Ours (pp) & 1024$\times$384 & \underline{0.109} & \underline{0.809} & \underline{4.705} & \textbf{0.193} & \underline{0.869} & \textbf{0.959} & \textbf{0.982} \\ \hline

\end{tabular}}

\label{tab:kitti_depth_unscaled}
\end{table*}

\subsection{KITTI Depth Prediction Benchmark}
We also measure the performance of our method on the KITTI Depth Prediction Benchmark using the 
metrics from~\cite{Uhrig2017THREEDV}. We train our method with a ResNet50 encoder on an image size of $1024\times320$ for $30$ epochs, and evaluate it using the online KITTI benchmark server.\footnote{\url{http://www.cvlibs.net/datasets/kitti/eval_depth.php?benchmark=depth_prediction}. See results under \texttt{g2s}} 

Results, ordered based on their rank, are shown in Table \ref{tab:kitti_depth_online} (D, M, and S represent supervised training with ground truth depths, monocular sequences, and stereo pairs, respectively. Seg represents additional supervised semantic segmentation training. G represents the use of GPS for multi-modal self-supervision).  We outperform all self-supervised methods while also performing better than many supervised methods which use ground truth depth maps during training.

\begin{table}[b]
\centering
\caption{Quantitative comparison on the KITTI Depth Prediction Benchmark (online server). }
\resizebox{\columnwidth}{!}{
\begin{tabular}{|l|c|c|c|c|c|}
\hline 
\textbf{Method}                 & \textbf{Train}            & \cellcolor{red!25}\textbf{SILog} & \cellcolor{red!25}\textbf{SqErrRel} & \cellcolor{red!25}\textbf{AbsErrRel} & \cellcolor{red!25}\textbf{iRMSE} \\ \hline \hline
DORN       ~\cite{fu2018deep}                   & D & 11.77 & 2.23 & 8.78 & 12.98 \\
SORD       ~\cite{diaz2019soft}                 & D & 12.39 & 2.49 & 10.10 & 13.48 \\
VNL        ~\cite{yin2019enforcing}             & D & 12.65 & 2.46 & 10.15 & 13.02 \\
DS-SIDENet ~\cite{ren2019deep}                  & D & 12.86 & 2.87 & 10.03 & 14.40 \\

PAP        ~\cite{zhang2019pattern}             & D & 13.08 & 2.72 & 10.27 & 13.95 \\
Guo \etal  ~\cite{guo2018learning}              & D+S & 13.41 & 2.86 & 10.60 & 15.06 \\
\hline \hline
Ours                                            & M+G & 14.16 & 3.65 & 11.40 & 15.53 \\
\hline \hline
Monodepth2 ~\cite{godard2019digging}            & M+S & 14.41 & 3.67 & 11.22 & 14.73 \\
DABC       ~\cite{li2018deep}                   & D & 14.49 & 4.08 & 12.72 & 15.53 \\
SDNet      ~\cite{ochs2019sdnet}                & D & 14.68 & 3.90 & 12.31 & 15.96 \\
APMoE      ~\cite{kong2019pixel}                & D & 14.74 & 3.88 & 11.74 & 15.63 \\
CSWS       ~\cite{li2018monocular}              & D & 14.85 & 348 & 11.84 & 16.38 \\

HBC        ~\cite{jiang2019hierarchical}        & D & 15.18 & 3.79 & 12.33 & 17.86 \\
SGDepth    ~\cite{klingner2020selfsupervised}   & M+Seg & 15.30 & 5.00 & 13.29 & 15.80 \\
DHGRL      ~\cite{zhang2018deep}                & D & 15.47 & 4.04 & 12.52 & 15.72 \\
MultiDepth ~\cite{liebel2019multidepth}         & D & 16.05 & 3.89 & 13.82 & 18.21 \\
LSIM       ~\cite{goldman2019learn}             & S & 17.92 & 6.88 & 14.04 & 17.62 \\
Monodepth  ~\cite{godard2017unsupervised}       & S & 22.02 & 20.58 & 17.79 & 21.84 \\ 
\hline
\end{tabular}}
\label{tab:kitti_depth_online}
\end{table}

\subsection{Ablation Studies}
To study the efficacy of the proposed \textit{g2s} loss in detail, we perform ablation studies on the introduced weighting strategy, as well as the frequency and dimensionality of the GPS used in the multi-modal training. 

\subsubsection{Weighting Strategy}
In Table \ref{tab:abl_depth}, we compare our proposed weighting strategy (Eq. \ref{eq:weighting}) against the alternative constant and linearly increasing weights for the \textit{g2s} loss.  The mean and standard deviation of scale factors as well as the corresponding metrics on \textit{scaled} predictions are shown. We confirm that utilizing an exponential weighting strategy effectively leverages the scale signal to produce scale-consistent and -aware depth estimates. As explained earlier in Section \ref{sec:method_weighting_strategy}, this is because penalizing the networks for the incorrect scales during the early training can interfere with the learning. Therefore, providing an increasing scale signal over the epochs, while allowing effective appearance-based learning in the early training, leads to the best results.


\begin{table}[hb]
\centering
\caption{
Scale factors on
Out-of-Distribution datasets.
}
\resizebox{0.65\columnwidth}{!}{
\begin{tabular}{|l|l|c|c|}
\hline
&\textbf{}\textbf{Method} & $\bf \mu_{scale}$ & \cellcolor{red!25}$\bf \sigma_{scale}\downarrow$  \\ \hline \hline
\multirow{3}{*}{\rotatebox[origin=c]{90}{M3D}} & SC-SfMLearner & 40.62 & 17.24 \\
& Monodepth2   & 76.02 & 24.40 \\
& Ours     & 2.81 & 0.85 \\ \hline
\multirow{3}{*}{\rotatebox[origin=c]{90}{CS}} & SC-SfMLearner  & 60.99 & 22.44 \\
& Monodepth2  & 118.61 & 36.49 \\
& Ours  & 4.01 & 1.22 \\ \hline
\end{tabular}}
\label{tab:scale_generalization}
\end{table}

\begin{table*}[bt]

\centering
\caption{Ablation study of different weighting strategies on the KITTI Eigen split~\cite{eigen2014depth}. }
\resizebox{0.75\textwidth}{!}{
\begin{tabular}{|l|c|c|cccc|ccc|}
\hline
\multirow{2}{*}{\textbf{Weights}} & \multirow{2}{*}{$\bf \mu_{scale}$} & \multirow{2}{*}{$\bf \sigma_{scale}\downarrow$} & \multicolumn{4}{c|}{\cellcolor{red!25}\textbf{Error$\downarrow$}} & \multicolumn{3}{c|}{\cellcolor{blue!25}\textbf{Accuracy$\uparrow$}} \\ \cline{4-10}
 &  & & Abs Rel & Sq Rel & RMSE & RMSE log & $\delta<1.25$ & $\delta<1.25^2$ & $\delta<1.25^3$ \\ \hline \hline
Const. $1$       & 0.776 & 0.126 & 1.280 & 53.454 & 21.915 & 0.934 & 0.217 & 0.427 & 0.604 \\ \hline
Const. $10^{-3}$ & 1.159 & 0.120 & 0.125 & 1.032 & 5.214 & 0.203 & 0.860 & 0.955 & 0.979 \\
Linear           & 0.124  & \textbf{0.020} & 0.443 & 4.757 & 12.083 & 0.588 & 0.303 & 0.561 & 0.766 \\
Ours (Eq. \ref{eq:weighting})            & \textbf{1.031} & \underline{0.073} & \textbf{0.112} & \textbf{0.894} & \textbf{4.852} &\textbf{ 0.192} & \textbf{0.877} & \textbf{0.958} & \textbf{0.981} \\ \hline
\end{tabular}}

\label{tab:abl_depth}
\end{table*}

\subsubsection{GPS Frequency and Dimensionality}
While GPS is ubiquitously copresent with driving video sequences, crowdsourced data often consists of high frames-per-second (fps) videos but lower frequency GPS. Furthermore, while altitude can be trilaterated by the GPS receivers, it is often not measured by the low-cost setups.  Therefore, we study the efficacy of the \textit{g2s} loss over different GPS frequencies, and the impact of the lack of altitude/height  measurements in two-dimensional GPS. 
\begin{figure}[t!]
\centering
  \includegraphics[width=0.7\linewidth]{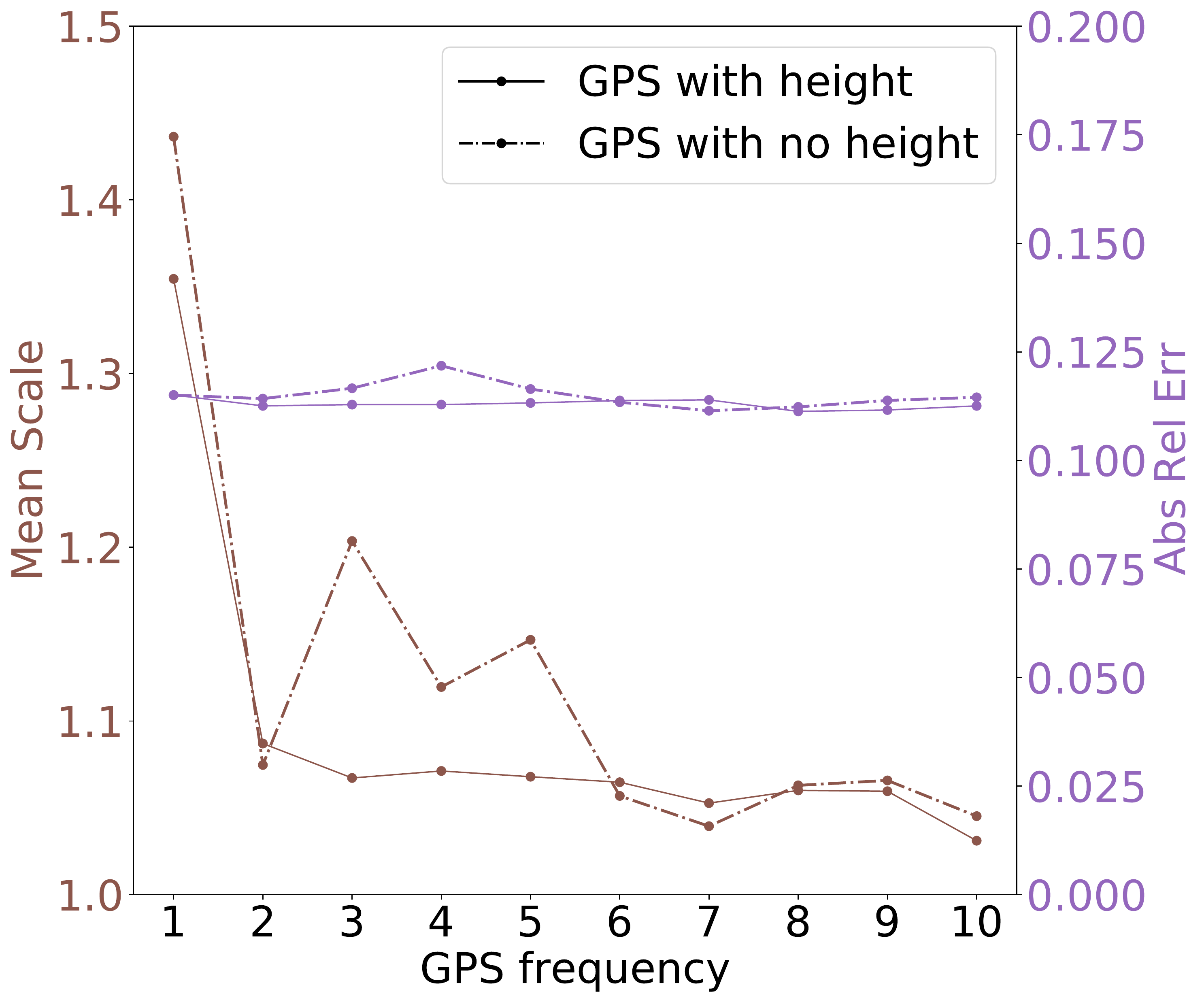}
  \caption{Ablation study on different GPS frequencies and dimensionality. Mean scale factor and performance  of depth estimation indicated by the Abs Rel Error~\cite{eigen2014depth} is shown.}
\label{fig:ablation_freq_gps}
\end{figure}

Note that the images in the KITTI dataset are captured at $10$ fps. To simulate the GPS frequencies lower than \SI{10}{\Hz} 
we randomly select the GPS data for $f<10$ frames for each $10$-frame non-overlapping subsequences ($\approx$ \SI{1}{s}) in the training data. Thereafter, we apply our \textit{g2s} loss as described in Eq. \ref{eq:g2s_loss} on the adjacent image pairs that have corresponding GPS available. The results are visualized in Fig. \ref{fig:ablation_freq_gps}.

We observe that our method is able to learn scale-aware depth estimation by using even the low-frequency GPS, thereby indicating the strength of the proposed \textit{g2s} loss. Our method improves upon the baseline Monodepth2~\cite{godard2019digging} even with a low-frequency scale-signal.
We also note that our method performs equally well without the availability of the altitude information. Thus, we conclude that our method would be applicable in the case of datasets with $2$-dimensional or sparse GPS.

\subsection{Out-of-Distribution Performance}
\label{sec:ood}
\begin{figure}[ht]
\centering
  \includegraphics[width=0.75\linewidth]{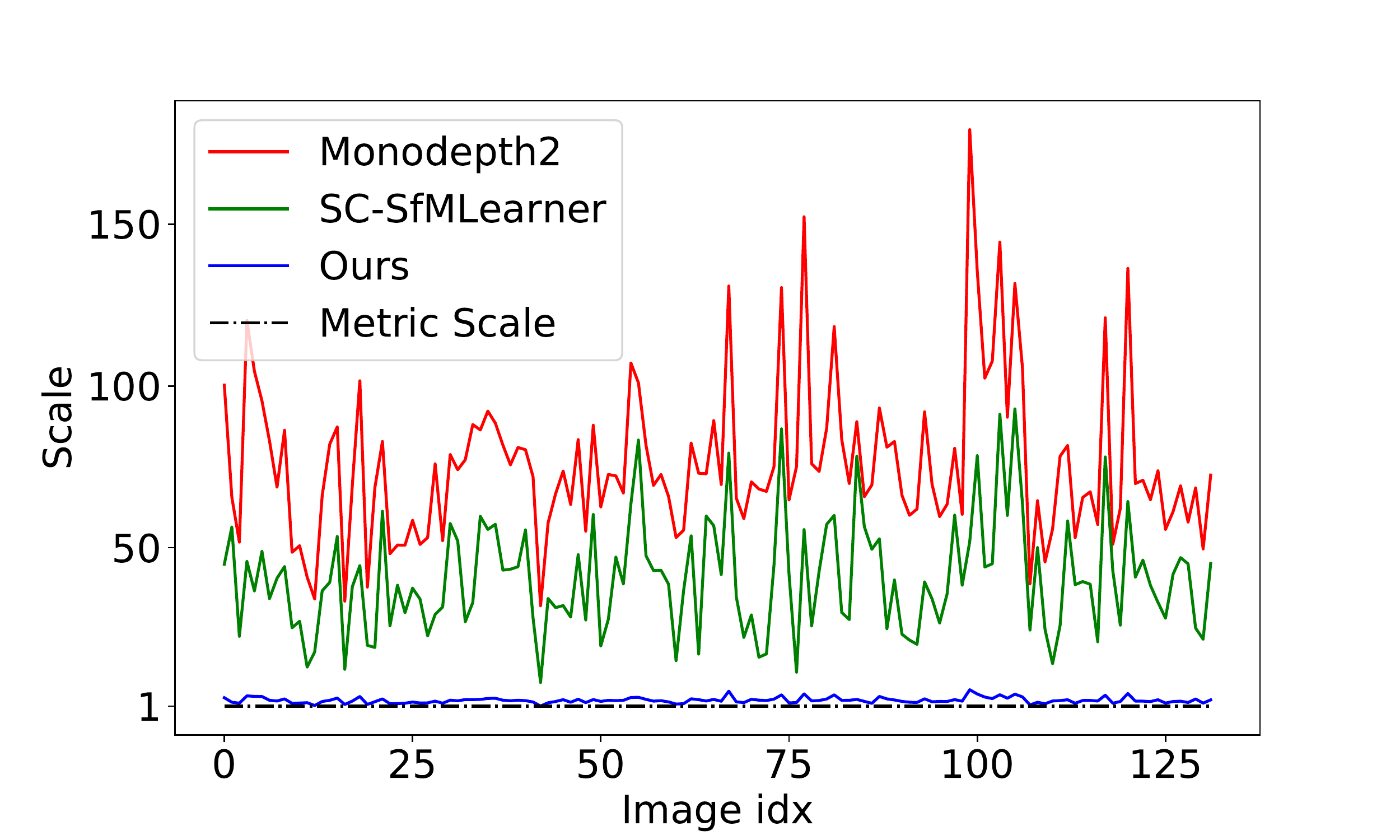}
  \caption{Out-of-Distribution depth scale variation on the Make3D test set~\cite{saxena2008make3d}. }
\label{fig:make3d_scales}
\end{figure}

\begin{figure}[ht]
\centering
  \includegraphics[width=0.75\columnwidth]{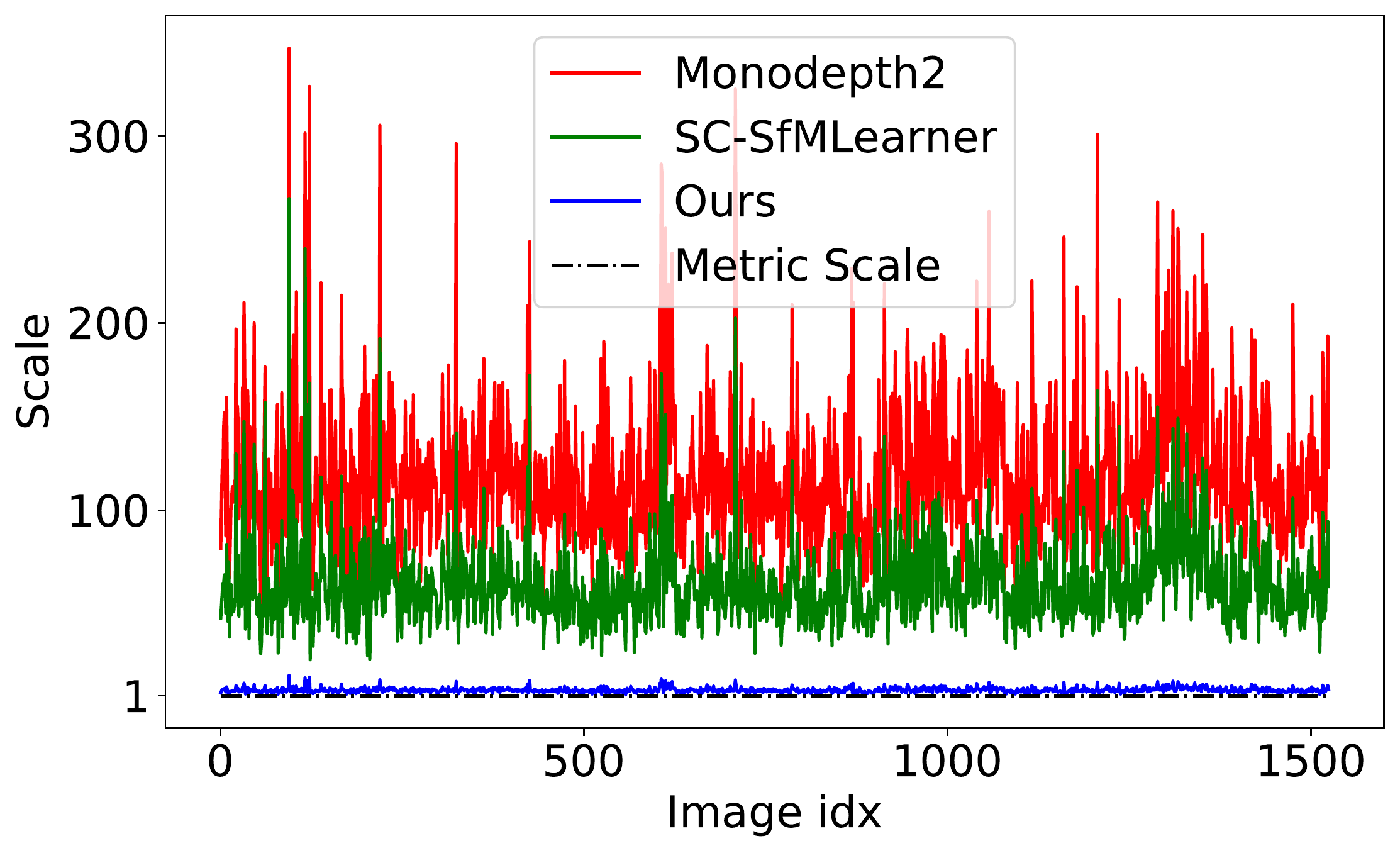}
  \caption{Out-of-Distribution depth scale variation on the Cityscapes test set~\cite{cordts2016cityscapes}. }
\label{fig:citys_scales}
\end{figure}

\begin{figure}[ht]
\centering
  \includegraphics[width=\linewidth]{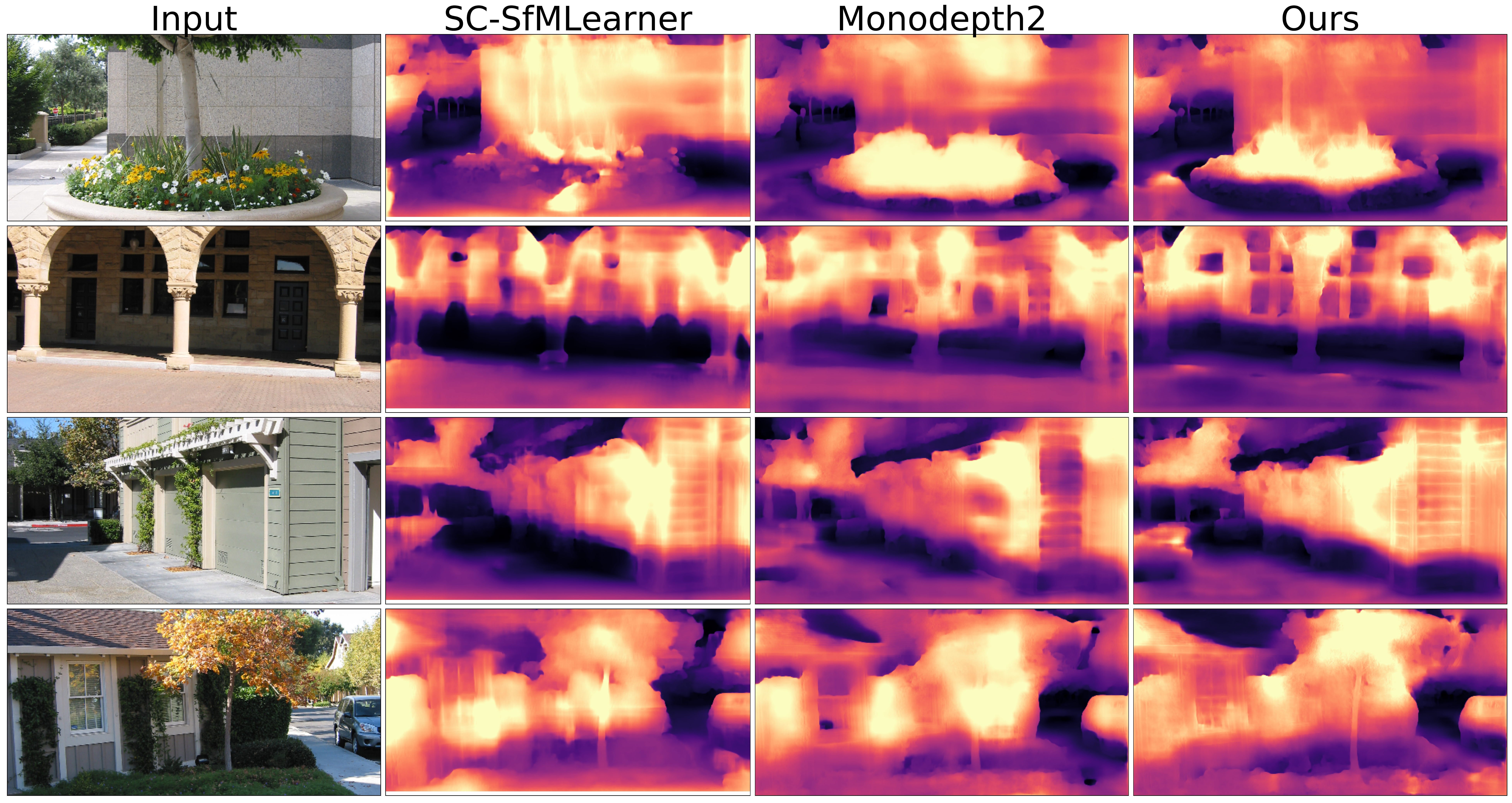}
  \caption{Qualitative results on Make3D test set~\cite{saxena2008make3d}. All methods were trained on the monocular sequences from the KITTI Eigen split~\cite{eigen2014depth}. Note that finer details are present in our predictions, such as building structures, silhouettes of flowers, and tree trunks.}
\label{fig:make3d_depth}
\end{figure}

\begin{figure}[hbt]
\centering
  \includegraphics[width=\linewidth]{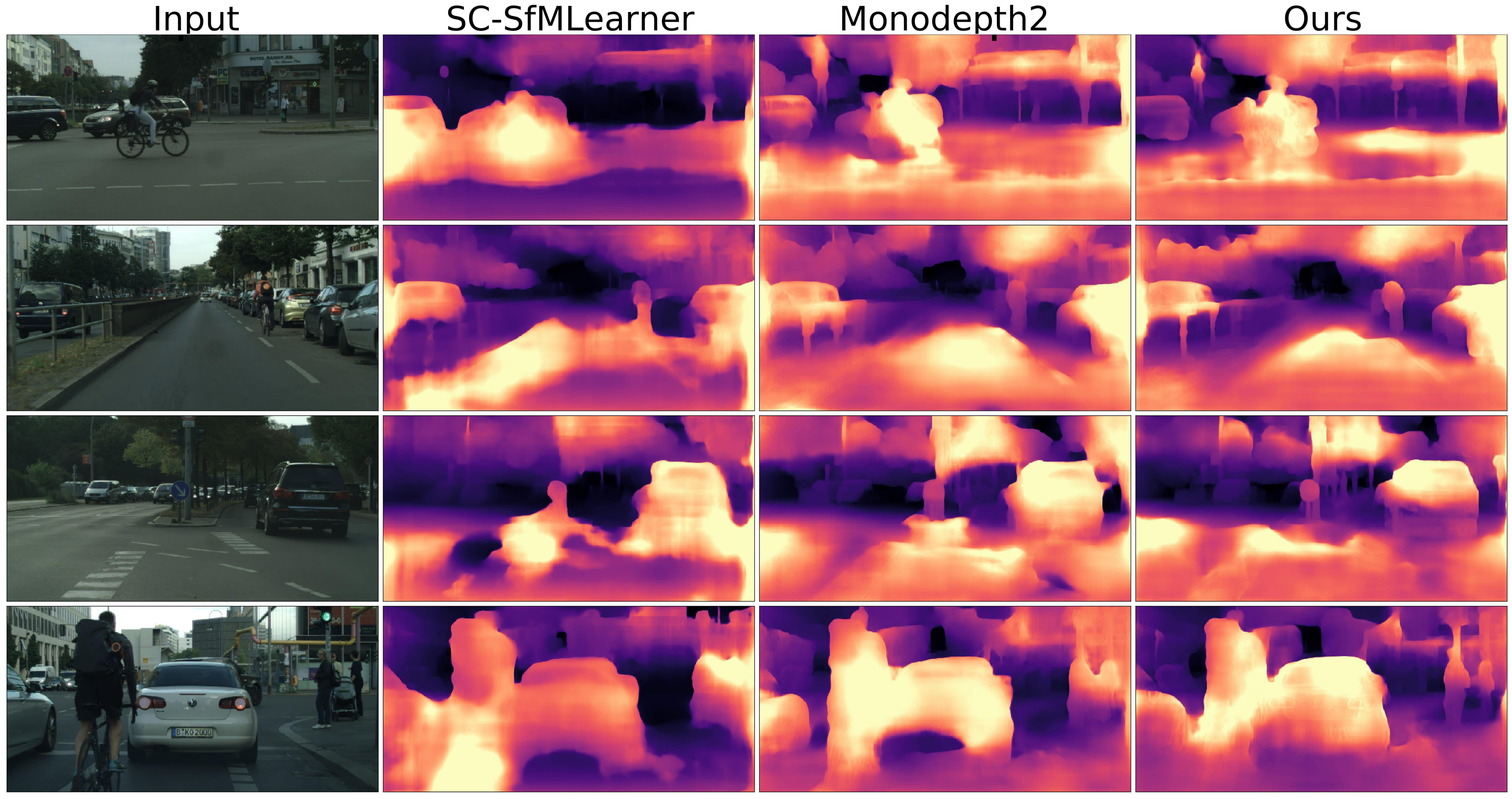}
  \caption{Qualitative results on Cityscapes test set~\cite{cordts2016cityscapes}. All methods were trained on the monocular sequences from the KITTI Eigen split. Note that finer details are present in our predictions, such as vehicle details, silhouettes of humans, and traffic signs.}
\label{fig:citys_depth}
\end{figure}

We also study the generalization capability of our method on o.o.d.~\cite{geirhos2020shortcut} datasets - Make3D (M3D)~\cite{saxena2008make3d} and Cityscapes (CS)~\cite{cordts2016cityscapes}. We evaluate our method (trained on the KITTI Eigen split) on the $2:1$ center crop of o.o.d. test images. 
Table \ref{tab:scale_generalization} shows the mean and standard deviation of the scale factors for the estimated depths, capped at \SI{70}{\m}. The standard deviation on the depth scale factor is the lowest for our method, indicating scale-consistency. This has also been visualized for the Make3D and Cityscapes test sets in Figs. \ref{fig:make3d_scales} and \ref{fig:citys_scales}. Also note that the mean of depth scale factors is significantly closer to $1$ than for other methods, even though metric-scale is no longer maintained. Finally, the qualitative results on the Make3D and Cityscapes dataset as shown in Figs. \ref{fig:make3d_depth} and \ref{fig:citys_depth},  demonstrate that the proposed multi-modal training improves the delineation of different objects in the depth estimation even for new scenes.  These results can be explained by the  learning of richer transferable discriminative features due to the scene and camera-setup independence of the GPS scale signal as explained in Sec \ref{sec:g2s_loss}.

\section{Conclusion}
\label{sec:conclusion}
This work addresses the problem of estimating scale-consistent and -aware monocular dense depths in a self-supervised setting, a feature essential for many practical autonomous vehicle applications. Previously, only appearance-based losses were used, and hence it was necessary to scale the predictions using the LiDAR ground truth. In contrast, by utilizing the camera-setup- and scene-independent GPS information, we propose an exponentially-weighted \textit{GPS-to-Scale (g2s)} loss to predict metrically accurate single-image depths within a multimodal self-supervised learning framework. No GPS information is used during the inference. Validating our approach on the KITTI dataset, we improve upon existing methods to predict sharper depths with finer-delineation of objects at scale. Through ablation studies, we also demonstrate the efficacy of our proposed loss, even when training on low-frequency or sparse GPS without height information. Finally, we show that our method results in better scale-consistency and -awareness even on out-of-distribution datasets. 
We posit that these improved results are a consequence of learning richer representations within a multimodal self-supervised framework. In the future, it seems promising to also study the impact of such at framework on the adversarial robustness of monocular depth estimation.




\bibliographystyle{IEEEtran}
\bibliography{IEEEabrv, ref.bib}

\addtolength{\textheight}{-12cm}   


\end{document}